\documentclass[sigconf,nonacm]{acmart}
\usepackage{subcaption}
\usepackage{xspace}
\usepackage{multirow}

\newcommand{\model}{EMODE\xspace}
\newcommand{\dpse}{DPSE\xspace}
\newcommand{\mepa}{MEPA\xspace}

\AtBeginDocument{%
  }

\copyrightyear{2026}
\acmYear{2026}
\setcopyright{cc}
\setcctype{by}
\acmConference[MM '26]{Proceedings of the 34th ACM International Conference on Multimedia}{November 10--14, 2026}{Rio de Janeiro, Brazil}
\acmBooktitle{Proceedings of the 34th ACM International Conference on Multimedia (MM '26), November 10--14, 2026, Rio de Janeiro, Brazil}
\acmDOI{10.1145/3767308.3836392}
\acmISBN{979-8-4007-2213-4/2026/11}

\begin{document}

\title{EMODE: Dynamic Para-Semantic Experts for Emotion-Aware Speech Language Modeling}
\titlenote{Accepted for ACM Multimedia 2026 (MM '26). DOI: \url{https://doi.org/10.1145/3767308.3836392}.}

\author{Jianan Pan}
\orcid{0009-0002-6774-1834}
\affiliation{%
  \institution{Zhejiang University}
  \city{Hangzhou}
  \country{China}}
\email{jonathan\_pan@zju.edu.cn}

\author{Yiwen Gu}
\orcid{0009-0006-5864-8397}
\affiliation{%
  \institution{Zhejiang University}
  \city{Hangzhou}
  \country{China}}
\email{yiwen\_gu@zju.edu.cn}

\author{Xinze Li}
\orcid{0009-0002-3311-8606}
\affiliation{%
  \institution{Zhejiang University}
  \city{Hangzhou}
  \country{China}}
\email{xinze.li.lab@zju.edu.cn}

\author{Rui Wang}
\orcid{0009-0009-0347-1215}
\affiliation{%
  \institution{Zhejiang University}
  \city{Hangzhou}
  \country{China}}
\email{22531069@zju.edu.cn}

\author{Kejie Huang}
\orcid{0000-0003-3722-9979}
\correspondingauthor
\affiliation{%
  \institution{Zhejiang University}
  \city{Hangzhou}
  \country{China}}
\email{huangkejie@zju.edu.cn}

\renewcommand{\shortauthors}{Jianan Pan, Yiwen Gu, Xinze Li, Rui Wang, \& Kejie Huang}
\begin{abstract}
Large speech language models have demonstrated strong capabilities in unified cross-modal understanding and generation, yet paralinguistic cues, especially emotion, remain difficult to preserve. Existing systems typically rely on entangled acoustic representations, which allow the underlying language model to depend excessively on recovered lexical content instead of grounding its behavior in acoustic-prosodic evidence. We address this limitation with EMODE, an emotion-aware speech language model built around \textbf{Dynamic Para-Semantic Experts (DPSE)}. DPSE decomposes continuous speech features into semantic and paralinguistic pathways, routes them dynamically, and fuses them before integration into the language model. To turn this structural decomposition into functional specialization, EMODE is trained with a three-stage curriculum consisting of semantic warm-up, paralinguistic activation, and joint refinement, guided by Orthogonal Expert Guidance (OEG), Semantic-to-Acoustic Alignment (SAA), and Gating Diversity Regularization (GDR). Experiments on SER test, empathetic response evaluation, and the newly constructed bilingual MEPA benchmark show that EMODE improves the balance between lexical fidelity and emotional sensitivity, strengthens affect-grounded response generation, and exposes the value of explicit para-semantic factorization for robust cross-corpus emotion understanding.
\end{abstract}


\maketitle

\section{Introduction}

Speech-language models are rapidly expanding the interface between audio and large language models. Whisper provides a strong foundation for robust large-scale speech encoding \cite{radford2022whisper}. SpeechT5 extends this direction to unified speech and text pretraining \cite{ao2022speecht5}. SpeechGPT shows that spoken interaction can be embedded directly into a conversational language-model pipeline \cite{zhang2023speechgpt}. More recent systems such as Qwen-Audio \cite{chu2023qwenaudio}, Qwen2-Audio \cite{chu2024qwen2audio}, SALMONN \cite{tang2023salmonn}, GLM-4-Voice \cite{zeng2024glm4voice}, Kimi-Audio \cite{kimiteam2025kimiaudio}, and Qwen2.5-Omni \cite{xu2025qwen25omni} further strengthen the case for unified spoken interaction. This trend is especially promising for human-centered applications because speech conveys not only lexical content but also timing, prosody, speaking style, and emotion.

Yet most speech-to-language-model connectors are still optimized primarily for semantic transduction. They are designed to preserve \emph{what} was said, while affective information is expected to survive implicitly inside a single acoustic representation. That bias is poorly matched to emotion-aware spoken interaction. Classical speech emotion research has long shown that emotion is distributed across prosody, voice quality, and temporal dynamics \cite{elayadi2011survey,abbaschian2021dlser}. Modern self-supervised encoders such as wav2vec~2.0 \cite{baevski2020wav2vec2}, HuBERT \cite{hsu2021hubert}, WavLM \cite{chen2022wavlm}, BEATs \cite{chen2023beats}, and emotion2vec \cite{ma2024emotion2vec} show that such cues can be learned at scale. Lightweight speech emotion recognition systems built on frozen speech encoders already perform competitively \cite{pepino2021wav2vec,morais2022serssl}. An emotion-aware speech connector therefore faces two coupled but non-identical requirements: it must preserve lexical alignment for language modeling, and it must retain paralinguistic evidence for emotion understanding.

Recent empathetic speech-language systems begin to address this gap by introducing emotion-aware supervision or empathy-oriented response objectives \cite{wang2024blspemo,wang2025opens2s,geng2025osumechat}. At the same time, recent speech emotion systems increasingly move beyond closed-set classification toward explanation or free-form description, as illustrated by SECap \cite{xu2024secap} and EmotionThinker \cite{wang2026emotionthinker}. Even so, most existing systems still rely on a single speech pathway, which leaves the model free to solve emotion-sensitive tasks through entangled representations or lexical shortcuts.

This work posits that emotion-aware speech-language modeling benefits from an explicit inductive bias that separates semantic and paralinguistic processing prior to generation. We instantiate this principle in \textbf{EMODE}, driven by its core module: \textbf{Dynamic Para-Semantic Experts (DPSE)}. DPSE routes acoustic features from a Whisper encoder into specialized semantic and paralinguistic experts, dynamically fusing their representations before feeding them to a Qwen language model. To alleviate the scarcity of natural, affect-sensitive dialog data, EMODE uses a frozen \emph{text teacher pathway} alongside emotion-aware continuation targets, drawing supervision from transcript-aligned and emotion-labeled resources. This architecture is optimized via a three-stage curriculum incorporating Orthogonal Expert Guidance (OEG), Semantic-to-Acoustic Alignment (SAA), and Gating Diversity Regularization (GDR), culminating in joint refinement of affective specialization and response generation.

\begin{figure}[t]
\centering
\includegraphics[width=\linewidth]{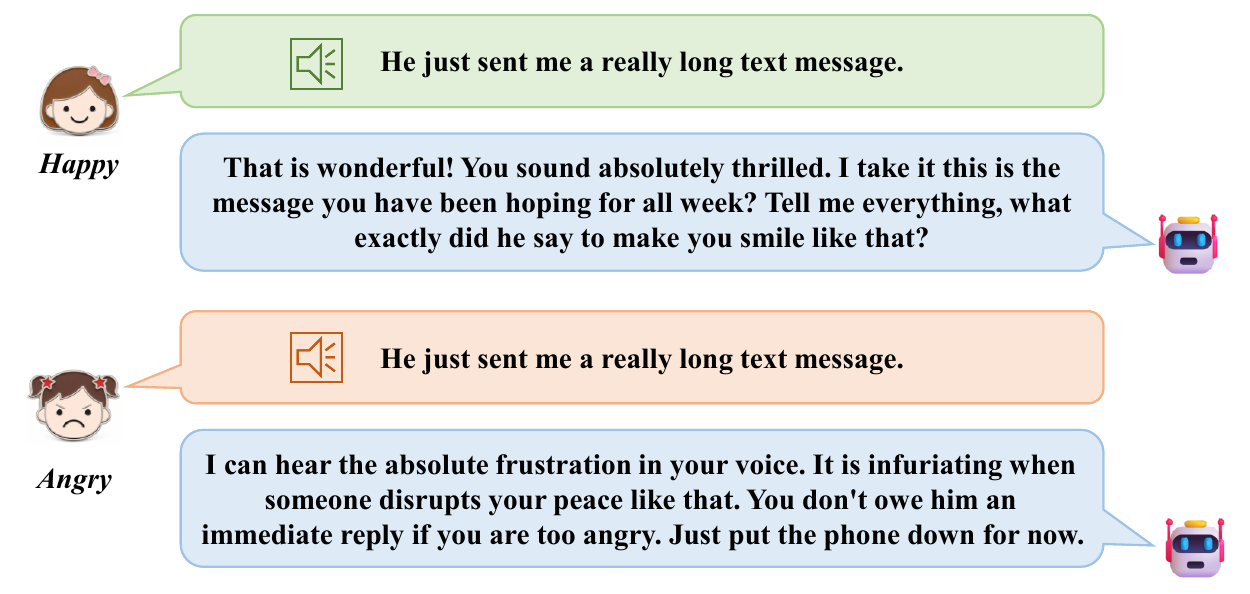}
\caption{Semantic-acoustic variation in human-machine interaction. Identical textual utterances can convey different intents depending on acoustic prosody. This example illustrates why end-to-end speech models should preserve and condition on paralinguistic evidence when generating empathetic responses.}
\Description{A single-column figure illustrating that matched lexical content with different emotional speaking styles can lead to different emotion-aware interpretations.}
\label{fig:dialogue_demo}
\end{figure}

We conduct a comprehensive evaluation of EMODE across standard in-domain \emph{speech emotion recognition (SER)} benchmarks (IEMOCAP and MELD) and diverse zero-shot datasets spanning five languages (RAVDESS, SAVEE, AESDD, MERBench, RESD, and CaFE). To test acoustic grounding while controlling lexical content, we introduce the \textbf{Multi-Emotion Parallel Audio (MEPA)} benchmark. Unlike conventional corpora that may contain transcript-level emotional regularities, MEPA groups diverse emotional speaking styles around fixed lexical statements across two languages, making acoustic-prosodic evidence necessary for distinguishing the variants within each group.

Our core contributions are threefold:
\begin{itemize}
\item We propose \textbf{EMODE}, an emotion-aware speech-language model featuring \textbf{DPSE}, which separates and dynamically recombines semantic and paralinguistic processing streams.
\item We introduce a coordinated three-stage training curriculum that translates structural decomposition into functional specialization through semantic and paralinguistic guidance, teacher-based alignment, and routing regularization.
\item We evaluate EMODE on in-domain and cross-lingual zero-shot SER, SpeechAlpaca empathetic response generation, and the new \mepa benchmark, covering both task-level performance and controlled lexical grounding.
\end{itemize}

\section{Related Work}
\label{sec:related_work}

\subsection{Speech-Language and Audio-Language Modeling}

Research on speech-conditioned language modeling has predominantly evolved along two parallel trajectories. The first focuses on developing robust speech representations and tighter speech-text interfaces. Foundational works such as Whisper \cite{radford2022whisper} provide robust large-scale speech encoding, while SpeechT5 \cite{ao2022speecht5} and SpeechGPT \cite{zhang2023speechgpt} pioneer unified speech-text pre-training and the direct integration of spoken interaction into conversational language models, respectively. Collectively, these systems establish the architectural backbone for practical, speech-native interactions.

The second trajectory broadens this interface to encompass general audio understanding and end-to-end multimodal dialogue. Models like Qwen-Audio and Qwen2-Audio \cite{chu2023qwenaudio,chu2024qwen2audio} unify audio instruction following, whereas SALMONN \cite{tang2023salmonn} and Pengi \cite{deshmukh2023pengi} extend capabilities to generic hearing and text-centric audio task formulation. Recent ``omni'' systems, including GLM-4-Voice \cite{zeng2024glm4voice}, Kimi-Audio \cite{kimiteam2025kimiaudio}, Phi-4-Multimodal \cite{microsoft2025phi4multimodal}, Megrez-3B-Omni \cite{li2025megrezomni}, MiniCPM-O \cite{openbmb2025minicpmo}, and Qwen2.5-Omni \cite{xu2025qwen25omni}, further push the boundaries of unified multimodal dialogue. 

\subsection{Speech Emotion Recognition and Emotion-Aware Spoken Interaction}

The field of Speech Emotion Recognition (SER) provides both the foundational acoustic cues and the rigorous evaluation paradigms relevant to our objective. Extensive surveys consistently identify prosody, timing, and voice quality as the primary carriers of affective information \cite{elayadi2011survey,abbaschian2021dlser}. Consequently, widely adopted corpora capture varied communicative contexts: IEMOCAP covers dyadic interactions \cite{busso2008iemocap}, MELD introduces multi-party dialogue \cite{poria2019meld}, RAVDESS and SAVEE offer controlled, acted speech \cite{livingstone2018ravdess,jackson2011savee}, and MSP-Podcast scales to naturalistic, in-the-wild conversational audio \cite{busso2025msppodcast}. Concurrently, the advent of self-supervised speech encoders---including wav2vec~2.0, HuBERT, WavLM, BEATs, and emotion2vec \cite{baevski2020wav2vec2,hsu2021hubert,chen2022wavlm,chen2023beats,ma2024emotion2vec}---has demonstrated that large-scale representations transfer effectively to downstream emotion understanding tasks.

Recent advancements increasingly integrate SER into broader conversational frameworks. Systems such as BLSP-Emo, OpenS2S, and OSUM-EChat inject explicit emotion-aware supervision into spoken dialogue agents \cite{wang2024blspemo,wang2025opens2s,geng2025osumechat}. Furthermore, SECap \cite{xu2024secap} and EmotionThinker \cite{wang2026emotionthinker} move beyond closed-set classification by generating descriptive captions and reasoning over emotional speech. \model complements these efforts. In addition to emotion-aware supervision and generation objectives, it addresses the preceding representation bottleneck by organizing semantic and paralinguistic evidence before integration into the language model.

\subsection{Expert Routing and Disentangled Speech Representations}

Sparse expert routing provides a robust mechanism for conditional computation. Sparsely gated Mixture-of-Experts (MoE) layers pioneered the dynamic activation of specialized subnetworks based on input characteristics \cite{shazeer2017outrageously}. Subsequent innovations, such as Switch Transformers \cite{fedus2022switch} and expert-choice routing \cite{zhou2022expertchoice}, refined large-scale allocation and capacity balancing. While \dpse adopts this principle of conditional specialization, it diverges significantly from traditional MoE applications: it employs precisely two experts and routes continuous speech features under explicit emotion-aware supervision, prioritizing feature disentanglement over language model capacity expansion.

This architectural choice fundamentally aligns \dpse with the literature on speech disentanglement. Prior works such as AutoVC \cite{qian2019autovc} factorize speaker identity from linguistic content, Triple Information Bottleneck models \cite{qian2020tripleib} decompose speech into multiple latent variables, and SpeechSplit2.0 \cite{chan2022speechsplit2} leverages unsupervised disentanglement for controllable synthesis. \model adapts this intuition for speech-language modeling. However, our primary objective is not synthesis control; rather, we aim to construct a speech representation that independently preserves semantic fidelity and affective faithfulness, ensuring both remain highly accessible for downstream language generation tasks.

\section{Approach}
\label{sec:approach}

Developing emotion-aware spoken interaction systems necessitates speech representations capable of capturing both lexical content (\emph{what} was said) and paralinguistic delivery (\emph{how} it was said). Fulfilling this dual objective is challenging when utilizing a single, entangled speech connector, particularly given the scarcity of naturally occurring speech-response data with explicit affective annotations. Building upon the teacher-guided alignment and emotion-aware continuation paradigms established in BLSP and BLSP-Emo \cite{wang2024blsp,wang2024blspemo}, \model leverages a frozen text teacher to supply response-side semantic targets and reformulates emotion-labeled speech into continuation-style supervision. These complementary techniques mitigate the data bottleneck by reducing reliance on massive datasets of directly annotated, affect-sensitive dialog pairs. Beyond this supervision paradigm, \model incorporates \textbf{Dynamic Para-Semantic Experts (DPSE)}, an architectural module that explicitly decouples semantic and paralinguistic processing pathways prior to representation fusion.

\begin{figure*}[t]
\centering
\includegraphics[width=\textwidth]{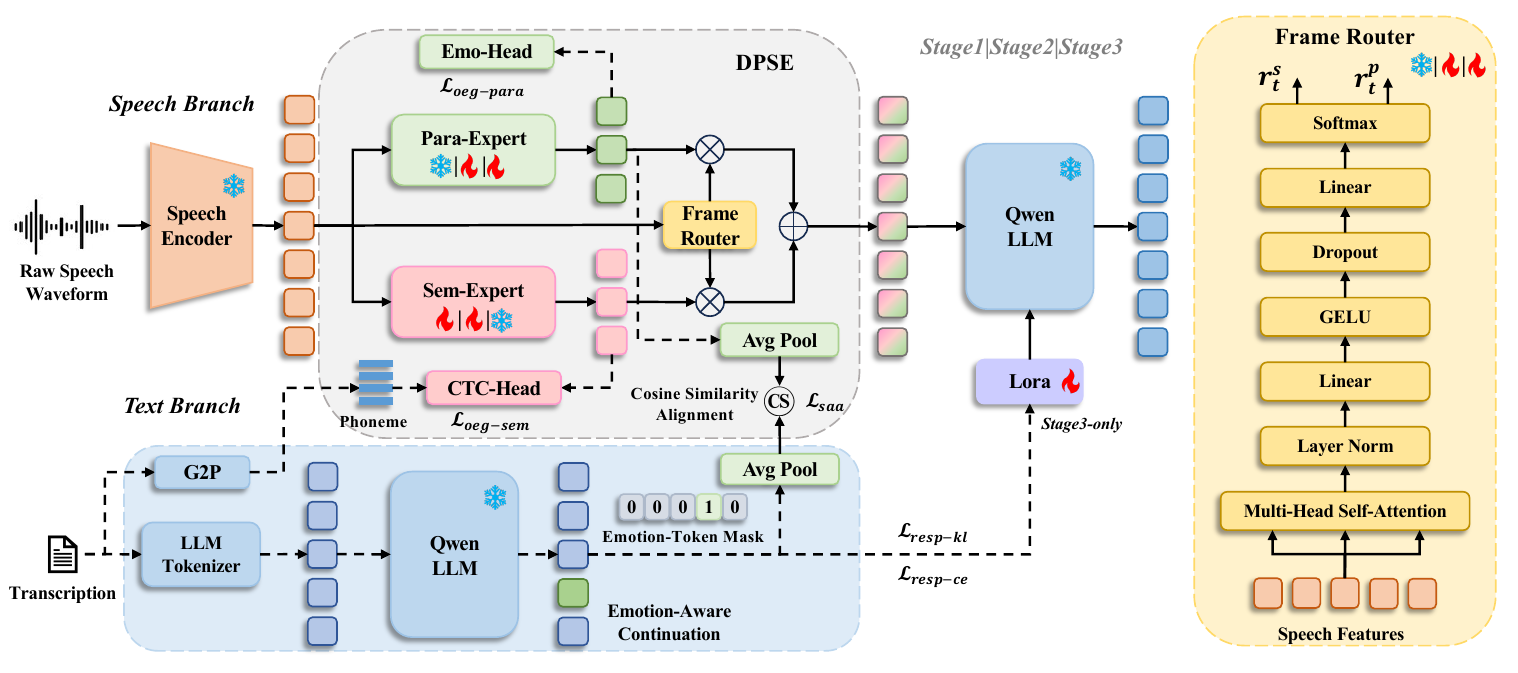}
\caption{Overview of EMODE. Whisper encodes the speech signal into frame-level acoustic features. DPSE routes these features into semantic and paralinguistic experts, fuses the two streams into a shared speech representation, and passes the result to the language model. A frozen text teacher pathway supplies response distributions and hidden states for teacher distillation and semantic-to-acoustic alignment during training. Grapheme-to-phoneme conversion produces phoneme targets for semantic CTC supervision, whereas the language model tokenizer produces the teacher-side token sequence used for response distillation and emotion-token masking.}
\Description{A double-column figure showing the EMODE architecture. Speech is encoded by Whisper, routed into semantic and paralinguistic experts, fused into a shared speech representation, and paired with a frozen text teacher pathway for distillation and alignment.}
\label{fig:method_overview}
\end{figure*}

\subsection{Task Setup and Dual-Path Supervision}
\label{subsec:problem_formulation}

Let $s$ denote a speech waveform, accompanied by an optional transcript $x = (x_1,\ldots,x_N)$, a target text response $y = (y_1,\ldots,y_M)$, and an emotion label $e$ (when available). The primary objective is to train a speech-language model that maintains strict lexical fidelity while preserving emotion-salient prosodic, temporal, and voice-quality cues.

The Whisper encoder first extracts frame-level acoustic features,
\begin{equation}
\mathbf{A} = f_{\mathrm{enc}}(s) \in \mathbb{R}^{T \times d_a}.
\end{equation}
DPSE then projects $\mathbf{A}$ into a downsampled sequence
\begin{equation}
\mathbf{H} = (\mathbf{h}_1,\ldots,\mathbf{h}_{T'}) \in \mathbb{R}^{T' \times d},
\end{equation}
which serves as the input to the language model backbone.

Training proceeds via two synchronized pathways processing the same sample. Let $c^{\mathrm{sp}}$ and $c^{\mathrm{txt}}$ represent the prompt tokens for the speech and text teacher pathways, respectively, and let $y_{<t} = (y_1,\ldots,y_{t-1})$ denote the teacher-forced response prefix at time step $t$. The speech pathway receives
\begin{equation}
\mathcal{I}^{\mathrm{sp}}_t = [\,c^{\mathrm{sp}};\mathbf{H};y_{<t}\,],
\end{equation}
while the text teacher pathway receives
\begin{equation}
\mathcal{I}^{\mathrm{txt}}_t = [\,c^{\mathrm{txt}};x;y_{<t}\,].
\end{equation}
The text teacher pathway operates in inference mode (without gradient updates) to generate reference token distributions and hidden states. When training on emotion-aware continuation data, the speech pathway receives no explicit text-based emotion marker. The target affect is therefore not disclosed through the prompt, encouraging the speech pathway to use acoustic evidence rather than a supplied emotion label. By combining transcript-aligned teacher targets with emotion-aware continuation supervision, this dual-path approach increases the usable supervision from paired speech-transcript and isolated emotion-labeled speech corpora despite the scarcity of natural, affect-grounded dialog data.

\subsection{Dynamic Para-Semantic Experts}
\label{subsec:dpse}

The DPSE module comprises three core components: a dynamic router, a semantic expert, and a paralinguistic expert. The router first contextualizes $\mathbf{A}$ with self-attention before predicting a frame-level, two-way routing distribution:
\begin{equation}
\mathbf{R} = \operatorname{Softmax}(f_{\mathrm{router}}(\mathbf{A})) \in [0,1]^{T \times 2},
\end{equation}
The two columns of $\mathbf{R}$ represent the routing weights for the semantic and paralinguistic pathways, respectively.

Both experts apply convolutional subsampling followed by lightweight residual processing to generate sequence representations:
\begin{equation}
\mathbf{S} = f_{\mathrm{sem}}(\mathbf{A}) \in \mathbb{R}^{T' \times d},
\qquad
\mathbf{P} = f_{\mathrm{para}}(\mathbf{A}) \in \mathbb{R}^{T' \times d}.
\end{equation}
To match the downsampled temporal resolution of the experts, the frame-level routing matrix $\mathbf{R}$ undergoes average pooling along the time axis to yield $\bar{\mathbf{R}} \in [0,1]^{T' \times 2}$. Letting $\mathbf{g} \in \mathbb{R}^{T' \times 1}$ denote the paralinguistic gating vector (extracted from the second column of $\bar{\mathbf{R}}$), the sequences are fused. The semantic stream serves as the backbone while the paralinguistic stream acts as a gated residual correction. The final fused representation is stabilized via layer normalization,
\begin{equation}
\mathbf{H} = \operatorname{LayerNorm}\!\big(\mathbf{S} + \mathbf{g} \odot \mathbf{P}\big),
\end{equation}
where $\odot$ denotes element-wise multiplication along the temporal dimension.

Additionally, the paralinguistic expert aggregates frame-level features via attention pooling to formulate an utterance-level summary,
\begin{equation}
\mathbf{u}^{(p)} = \sum_{t=1}^{T'} \alpha_t \mathbf{p}_t,
\qquad
\alpha_t = \frac{\exp(b_t)}{\sum_{u=1}^{T'} \exp(b_u)},
\end{equation}
which subsequently feeds into an auxiliary emotion classifier. Finally, the paralinguistic sequence is transformed via a linear projection $W_p$ to produce
\begin{equation}
\tilde{\mathbf{P}} = W_p \mathbf{P},
\end{equation}
providing the requisite representation for the semantic-to-acoustic alignment objective.

\subsection{Orthogonal Expert Guidance}
\label{subsec:dpse_training_strategy}

Introducing dual experts does not inherently guarantee functional specialization. Optimized solely under a response objective, both branches may converge toward identical lexical solutions, potentially causing router collapse before distinct roles emerge. \model mitigates this via \textbf{Orthogonal Expert Guidance (OEG)}, which imposes distinct supervisory signals on each expert.

Given transcript $x$, the proposed framework uses two parallel symbolic views. A grapheme-to-phoneme module $\gamma(\cdot)$ maps the transcript to a phoneme sequence
\begin{equation}
\gamma(x) = u = (u_1,\ldots,u_K),
\qquad
u_k \in \mathcal{V}_{\mathrm{ph}},
\end{equation}
where $\mathcal{V}_{\mathrm{ph}}$ denotes the phoneme vocabulary. In parallel, the language model tokenizer $\tau(\cdot)$ maps textual context to the teacher-side token sequence used by the frozen text pathway. These two views serve different objectives: $u$ supervises the semantic branch through phoneme-level CTC, whereas $\tau(\cdot)$ indexes teacher hidden states and token masks for response distillation and semantic-to-acoustic alignment. No positionwise correspondence between phoneme symbols and language model tokens is assumed.

For the semantic expert, OEG employs a phoneme-level Connectionist Temporal Classification (CTC) loss,
\begin{equation}
\mathcal{L}_{\mathrm{oeg\text{-}sem}}
=
\operatorname{CTC}\!\big(W_{\mathrm{ctc}}\mathbf{S}, u\big),
\end{equation}
where $W_{\mathrm{ctc}}\mathbf{S} \in \mathbb{R}^{T' \times |\mathcal{V}_{\mathrm{ph}}|}$ is the frame-level phoneme distribution produced by the CTC prediction head, $u=\gamma(x)$ is the phoneme target sequence, and $\operatorname{CTC}(\cdot)$ is the standard Connectionist Temporal Classification loss \cite{graves2006connectionist}. This phoneme-level targeting aligns with our final experimental training configuration.

Conversely, OEG regulates the paralinguistic expert using emotion supervision at the utterance level:
\begin{equation}
\hat{e}^{(p)} = g_{\mathrm{emo}}\!\big(\mathbf{u}^{(p)}\big),
\qquad
\mathcal{L}_{\mathrm{oeg\text{-}para}}
=
\operatorname{CrossEntropy}\!\big(\hat{e}^{(p)}, e\big),
\end{equation}
where $g_{\mathrm{emo}}(\cdot)$ denotes the emotion classifier and $\operatorname{CrossEntropy}(\cdot)$ represents the standard categorical cross-entropy function. Collectively, these orthogonal losses encourage lexical consistency in the semantic branch and affective sensitivity in the paralinguistic branch.

\subsection{Semantic-to-Acoustic Alignment}
\label{subsec:speech_language_objectives}

Despite OEG, the paralinguistic representations may remain weakly grounded in the task-relevant semantic context required by the language model. To bridge this gap, we introduce \textbf{Semantic-to-Acoustic Alignment (SAA)}. Let
\begin{equation}
\mathbf{Z} = (\mathbf{z}_1,\ldots,\mathbf{z}_L) \in \mathbb{R}^{L \times d}
\end{equation}
denote the hidden states extracted from the frozen text teacher pathway. We compute mask-aware, mean-pooled contextual summaries for both the speech and teacher modalities:
\begin{equation}
\mathbf{q}
=
\frac{\sum_{t=1}^{T'} m^{\mathrm{sp}}_t \tilde{\mathbf{p}}_t}{\sum_{t=1}^{T'} m^{\mathrm{sp}}_t},
\qquad
\mathbf{v}
=
\frac{\sum_{\ell=1}^{L} m^{\mathrm{txt}}_\ell \mathbf{z}_\ell}{\sum_{\ell=1}^{L} m^{\mathrm{txt}}_\ell},
\end{equation}
where $m^{\mathrm{sp}}$ represents the valid speech-frame mask, $m^{\mathrm{txt}}$ is the teacher-side token mask, and $\tilde{\mathbf{p}}_t$ denotes the projected paralinguistic feature at time step $t$. The alignment loss is formulated as:
\begin{equation}
\mathcal{L}_{\mathrm{saa}}
=
1 - \operatorname{cosine}\!\big(\mathbf{q}, \mathbf{v}\big),
\end{equation}
where $\operatorname{cosine}(\cdot,\cdot)$ denotes cosine similarity. Let
\begin{equation}
\tau([c^{\mathrm{txt}};x;y]) = (w_1,\ldots,w_L)
\end{equation}
denote the teacher-side token sequence. For a label-supervised SER sample with emotion label $e$, let $v(e)$ denote the associated textual verbalizer and let $\mathcal{E}$ denote the set of token positions occupied by that verbalizer within $\tau([c^{\mathrm{txt}};x;y])$. The emotion-token mask is then defined as
\begin{equation}
m^{\mathrm{txt}}_\ell = \mathbf{1}\!\left[\ell \in \mathcal{E}\right].
\end{equation}

\subsection{Gating Diversity Regularization}
\label{subsec:stagewise_optimization}

Without intervention, the router may prematurely overcommit to the faster-learning semantic pathway and hinder subsequent specialization. We therefore apply \textbf{Gating Diversity Regularization (GDR)} to the initial frame-level routing probabilities:
\begin{equation}
\mathcal{L}_{\mathrm{gdr}}
=
-\frac{1}{\sum_{t=1}^{T} m_t}
\sum_{t=1}^{T} m_t \, \mathcal{H}(\mathbf{r}_t),
\end{equation}
where $\mathcal{H}(\cdot)$ computes the entropy, $m_t$ is the valid-frame mask, and $\mathbf{r}_t$ represents the router distribution at frame $t$. Minimizing negative entropy promotes broad routing exploration early in training, while decay of its coefficient later permits more selective routing.

\subsection{Stage-Wise Optimization}
\label{subsec:stagewise_recipe}

The initial training stages leverage teacher distillation to stabilize speech-text alignment, whereas the final stage transitions to autoregressive response generation. Let $\mathcal{L}_{\mathrm{resp\text{-}kl}}$ denote the response-level Kullback-Leibler (KL) divergence loss employed for teacher distillation:
\begin{equation}
\mathcal{L}_{\mathrm{resp\text{-}kl}}
=
\operatorname{KL}\!\big(p^{\mathrm{txt}}_{\mathrm{resp}} \,\|\, p^{\mathrm{sp}}_{\mathrm{resp}}\big),
\end{equation}
where $p^{\mathrm{txt}}_{\mathrm{resp}}$ and $p^{\mathrm{sp}}_{\mathrm{resp}}$ represent the next-token probability distributions at the response positions for the teacher and speech pathways, respectively. Furthermore, let $\mathcal{L}_{\mathrm{resp\text{-}ce}}$ represent the standard autoregressive cross-entropy loss over response tokens.

\paragraph{Stage I: semantic warm-up.}
The paralinguistic expert remains frozen, dynamic routing is disabled (biasing the network entirely toward the semantic pathway), and optimization strictly prioritizes lexical faithfulness:
\begin{equation}
\mathcal{L}^{\mathrm{I}}
=
\mathcal{L}_{\mathrm{resp\text{-}kl}}
+
\lambda_{\mathrm{sem}} \mathcal{L}_{\mathrm{oeg\text{-}sem}}.
\end{equation}
This phase establishes a stable semantic interface between the acoustic encoder and the language model.

\paragraph{Stage II: paralinguistic activation.}
The paralinguistic expert is unfrozen, routing becomes fully dynamic, and the language model remains frozen. Under our standard training configuration, this stage minimizes:
\begin{equation}
\mathcal{L}^{\mathrm{II}}
=
\lambda_{\mathrm{sem}} \mathcal{L}_{\mathrm{oeg\text{-}sem}}
+
\lambda_{\mathrm{para}} \mathcal{L}_{\mathrm{oeg\text{-}para}}
+
\lambda_{\mathrm{gdr}} \mathcal{L}_{\mathrm{gdr}}.
\end{equation}
Stage II marks the critical juncture where the dual experts acquire their distinct functional roles, safeguarded by GDR against premature routing collapse.

\paragraph{Stage III: joint refinement.}
The culminating stage trains on a mixture of speech continuation, emotion-aware continuation, and label-supervised SER data. It introduces low-rank adaptations to the language model and optimizes the joint objective:
\begin{equation}
\mathcal{L}^{\mathrm{III}}
=
\mathcal{L}_{\mathrm{resp\text{-}ce}}
+
\lambda_{\mathrm{para}} \mathcal{L}_{\mathrm{oeg\text{-}para}}
+
\lambda_{\mathrm{saa}}(t) \mathcal{L}_{\mathrm{saa}}
+
\lambda_{\mathrm{gdr}}(t) \mathcal{L}_{\mathrm{gdr}}.
\end{equation}
Under the final configuration, the paralinguistic OEG loss weight is fixed at $0.02$, the SAA weight scales linearly from $0$ to $0.1$, and the GDR coefficient decays from $0.03$ to $0.01$, accommodating increasingly selective routing as training progresses.

In summary, \model synthesizes structured para-semantic factorization with robust teacher-guided and emotion-aware supervision. DPSE provides the foundational architectural bias, while the synergistic application of OEG, SAA, and GDR ensures this decomposition remains functionally meaningful throughout the training lifecycle.

\section{Experiments}
\label{sec:experiments}

\begin{table*}[!t]
\centering
\small
\caption{Speech emotion recognition accuracy (\%) on the main evaluation suite. IEMOCAP and MELD are in-domain. RAVDESS, SAVEE, MERBench Test1, MERBench Test2, RESD (Ru.), AESDD (Gr.), and CaFE (Fr.) are zero-shot evaluation sets; Ru., Gr., and Fr. denote Russian, Greek, and French test conditions. MERBench Test1 and Test2 are two evaluation splits from the Chinese MER2023 benchmark. The best and second-best entries in each column are marked in \textbf{bold} and \underline{underline}, respectively.}
\label{tab:main_results}
\renewcommand{\arraystretch}{1.08}
\setlength{\tabcolsep}{3.5pt}
\resizebox{\textwidth}{!}{%
\begin{tabular}{lcccccccccc}
\toprule
\multicolumn{1}{c}{\multirow{2}{*}{\textbf{Model}}} & \multicolumn{2}{c}{\textbf{In-domain}} & \multicolumn{7}{c}{\textbf{Zero-shot}} & \multirow{2}{*}{\textbf{Avg}} \\
\cmidrule(lr){2-3}\cmidrule(lr){4-10}
& \textbf{IEMOCAP} & \textbf{MELD} & \textbf{RAVDESS} & \textbf{SAVEE} & \textbf{MERBench Test1} & \textbf{MERBench Test2} & \textbf{RESD (Ru.)} & \textbf{AESDD (Gr.)} & \textbf{CaFE (Fr.)} & \\
\midrule
\multicolumn{11}{c}{\textit{General Speech LLMs}} \\
\midrule
GLM-4-Voice \cite{zeng2024glm4voice} & 22.38 & 21.43 & 19.67 & 20.84 & 17.91 & 16.48 & 19.05 & 18.52 & 18.73 & 19.45 \\
BLSP \cite{wang2024blsp} & 41.24 & 50.47 & 11.10 & 10.77 & 18.24 & 17.02 & 16.41 & 15.36 & 19.18 & 22.20 \\
DiVA \cite{held2024diva} & 40.15 & 35.19 & 18.28 & 20.14 & 22.35 & 21.08 & 20.41 & 19.66 & 22.17 & 24.38 \\
SALMONN \cite{tang2023salmonn} & 23.82 & 31.32 & 20.38 & 17.86 & 20.53 & 19.86 & 18.92 & 18.74 & 19.44 & 21.21 \\
Qwen2-Audio \cite{chu2024qwen2audio} & 37.71 & 51.23 & 64.98 & 65.13 & 61.75 & 59.42 & 58.31 & 57.84 & 62.48 & 57.65 \\
Kimi-Audio \cite{kimiteam2025kimiaudio} & 57.72 & \underline{59.13} & 61.07 & 55.21 & 56.92 & 55.38 & 54.76 & 53.84 & 58.43 & 56.94 \\
\midrule
\multicolumn{11}{c}{\textit{Emotion-Focused Speech LLMs}} \\
\midrule
SECap \cite{xu2024secap} & 36.29 & 34.20 & 28.34 & 21.93 & 26.88 & 25.47 & 23.95 & 24.16 & 26.14 & 27.48 \\
OSUM-EChat \cite{geng2025osumechat} & 41.49 & 53.38 & 37.34 & 24.27 & 35.67 & 33.92 & 31.58 & 31.84 & 34.41 & 35.99 \\
BLSP-Emo \cite{wang2024blspemo} & 76.00 & 57.30 & \underline{72.00} & 63.73 & 58.76 & 55.28 & \textbf{66.88} & 61.60 & \textbf{75.31} & 65.21 \\
\midrule
\multicolumn{11}{c}{\textit{Omni LLMs}} \\
\midrule
Phi-4-Multimodal \cite{microsoft2025phi4multimodal} & 36.62 & 39.81 & 17.63 & 15.55 & 20.11 & 19.37 & 18.64 & 18.92 & 19.82 & 22.94 \\
MiniCPM-O \cite{openbmb2025minicpmo} & 35.54 & 52.78 & 40.93 & 35.47 & 37.84 & 35.68 & 34.12 & 33.56 & 36.27 & 38.02 \\
Qwen2.5-Omni \cite{xu2025qwen25omni} & 45.70 & 54.64 & 64.77 & 52.49 & 53.61 & 51.24 & 48.93 & 49.83 & 52.17 & 52.60 \\
EmotionThinker \cite{wang2026emotionthinker} & \textbf{77.68} & 59.71 & 71.56 & \underline{73.96} & \underline{69.18} & \underline{62.47} & \underline{65.42} & \underline{63.42} & 72.74 & \underline{68.46} \\
\midrule
\textbf{EMODE} & \underline{76.68} & \textbf{64.23} & \textbf{75.58} & \textbf{85.76} & \textbf{75.14} & \textbf{66.94} & 64.56 & \textbf{66.57} & \underline{73.46} & \textbf{72.10} \\
\bottomrule
\end{tabular}}
\end{table*}

\subsection{Experimental Setup}
\label{subsec:experimental_setup}

\paragraph{Training corpora.}
Stage I uses speech continuation data from LibriSpeech \cite{panayotov2015librispeech}, Common Voice \cite{ardila2020commonvoice}, GigaSpeech \cite{chen2021gigaspeech}, and WenetSpeech \cite{zhang2022wenetspeech}. These corpora provide diverse speech-transcript pairs to stabilize semantic alignment prior to introducing emotion supervision. To overcome the scarcity of natural, affect-grounded speech-response data, Stages II and III incorporate emotion supervision from IEMOCAP \cite{busso2008iemocap}, MELD \cite{poria2019meld}, ESD \cite{zhou2022esd}, CMU-MOSEI \cite{zadeh2018mosei}, MEAD \cite{wang2020mead}, EARS \cite{richter2024ears}, EXPRESSO \cite{nguyen2023expresso}, and MSP-Podcast \cite{busso2025msppodcast}. All emotion datasets are mapped into a unified five-way label space: \emph{neutral}, \emph{happy}, \emph{angry}, \emph{sad}, and \emph{surprised}.

\paragraph{Evaluation corpora.}
Our primary evaluation suite comprises two in-domain corpora (IEMOCAP and MELD) and seven zero-shot test sets: RAVDESS \cite{livingstone2018ravdess}, SAVEE \cite{jackson2011savee}, MERBench Test1 and Test2 \cite{lian2024merbench}, RESD \cite{aniemore2022resd}, AESDD \cite{vryzas2018aesdd}, and CaFE \cite{gournay2018cafe}. This diverse selection extends our evaluation beyond English to encompass Greek, Chinese, Russian, and French conditions. For spoken empathetic interaction, we evaluate on SpeechAlpaca using the speech-only protocol established by BLSP-Emo \cite{wang2024blspemo}. We also introduce \mepa, a bilingual synthetic benchmark built from lexically shared first-person statements and emotion-specific empathetic responses rendered by expressive speech models. Within each group, the statement remains fixed while the speaking style varies across emotions. This controlled structure reduces transcript-level shortcuts and complements conventional corpora by testing whether predictions remain sensitive to acoustic-prosodic variation.

\paragraph{Model and training overview.}
\model employs Whisper-base as the speech encoder, \dpse as the bridging connector, and Qwen2.5-7B-Instruct as the language model backbone. The semantic branch receives phoneme-level CTC supervision, while the paralinguistic branch is guided by utterance-level emotion labels. Their fused representation is subsequently refined via routed para-semantic residual fusion. Unless otherwise specified, all reported results reflect the default three-stage training curriculum detailed in Section~\ref{sec:approach}: semantic warm-up, paralinguistic activation, and joint refinement.

\subsection{Tasks and Metrics}
\label{subsec:tasks_metrics}

\paragraph{Speech Emotion Recognition (SER)}
We formulate SER as a closed-set, audio-conditioned classification task executed via text generation. Given an input utterance and a dataset-specific label inventory (a subset of the unified five-way training space), \model is prompted to generate a single emotion label. The generated outputs are normalized to canonical labels via alias matching and evaluated using standard classification accuracy.

\paragraph{Empathetic Response.}
The SpeechAlpaca benchmark evaluates a model's ability to generate helpful and affectively appropriate responses from speech input. The primary metrics are \emph{quality}, reflecting helpfulness and coherence, and \emph{empathy}, measuring how well a response acknowledges the speaker's emotional state. Both metrics are scored on a $[0,10]$ scale by a unified DeepSeek-based evaluator across all methods. SpeechAlpaca also includes a four-way speech-only SER protocol over \emph{neutral}, \emph{happy}, \emph{angry}, and \emph{sad}; we report this score as an auxiliary measure rather than the primary response criterion. These single-judge scores are not a substitute for human evaluation.

\paragraph{MEPA benchmark.}
While \mepa shares the audio-only label-generation protocol of standard SER tasks, its structural design exposes models that still rely on lexical shortcuts. Each evaluation group contains one lexical statement rendered in five distinct emotional styles, so accurate prediction must depend on acoustic evidence rather than transcript priors. We therefore report sample-level accuracy, macro-F1, and \emph{Group-Level Complete Accuracy} (GLCA),
\begin{equation}
\mathrm{GLCA}
=
\frac{1}{|\mathcal{G}|}
\sum_{g \in \mathcal{G}}
\mathbf{1}\!\left[
\forall (x_i, y_i)\in g,\ \hat{y}_i = y_i
\right],
\end{equation}
where $\mathcal{G}$ denotes the set of lexical groups, each containing all five emotional realizations of one shared statement. Because every evaluated group is complete, GLCA measures the fraction of statements for which the model correctly identifies all five realizations, making it a stricter complement to sample-level accuracy and macro-F1.

\begin{figure*}[!t]
\centering
\IfFileExists{figs/mepa_en_norm.png}{%
    \begin{subfigure}[t]{0.47\textwidth}
    \centering
    \includegraphics[height=2.3in,keepaspectratio]{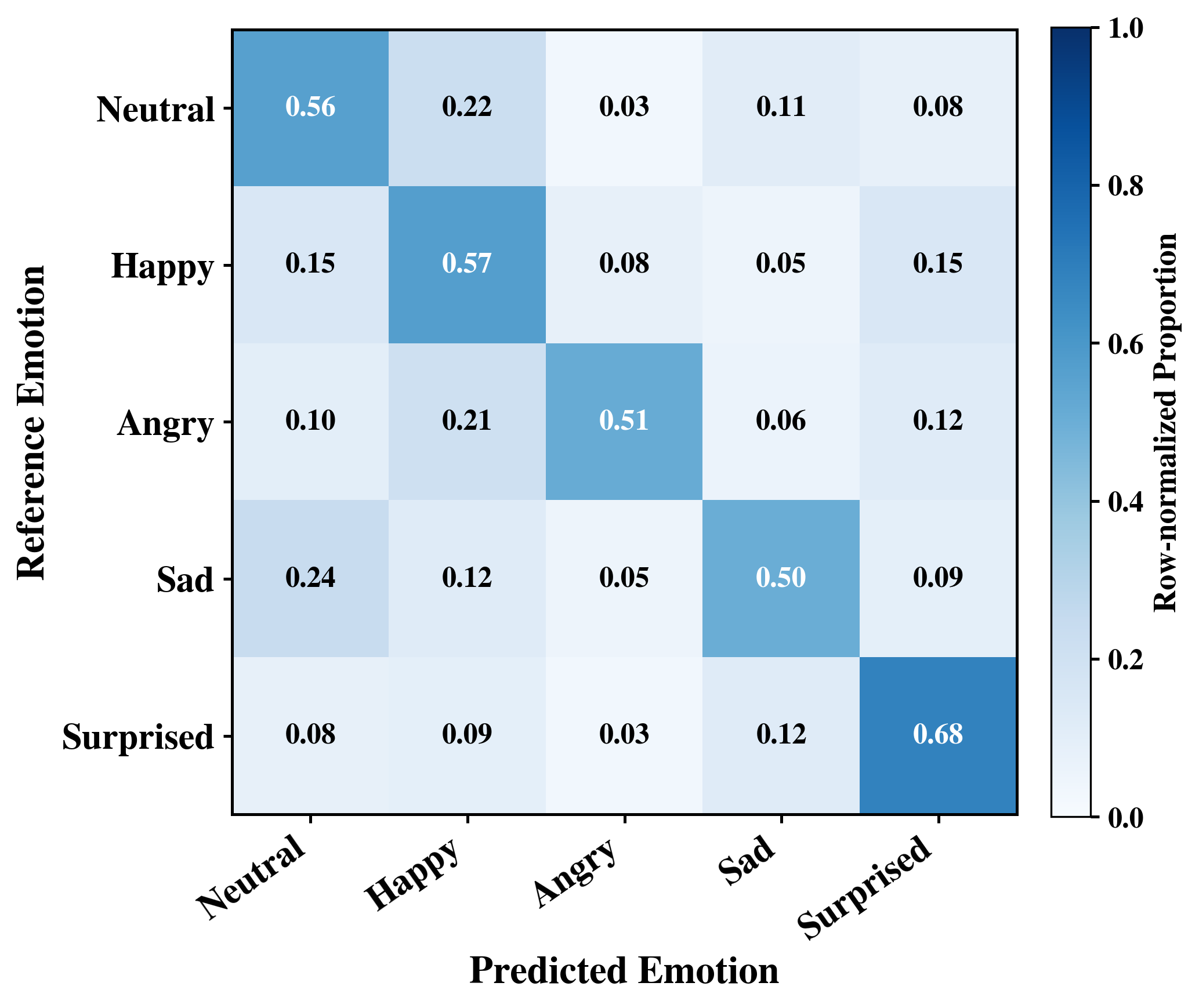}
    \caption{English subset.}
    \end{subfigure}\hfill
    \begin{subfigure}[t]{0.47\textwidth}
    \centering
    \includegraphics[height=2.3in,keepaspectratio]{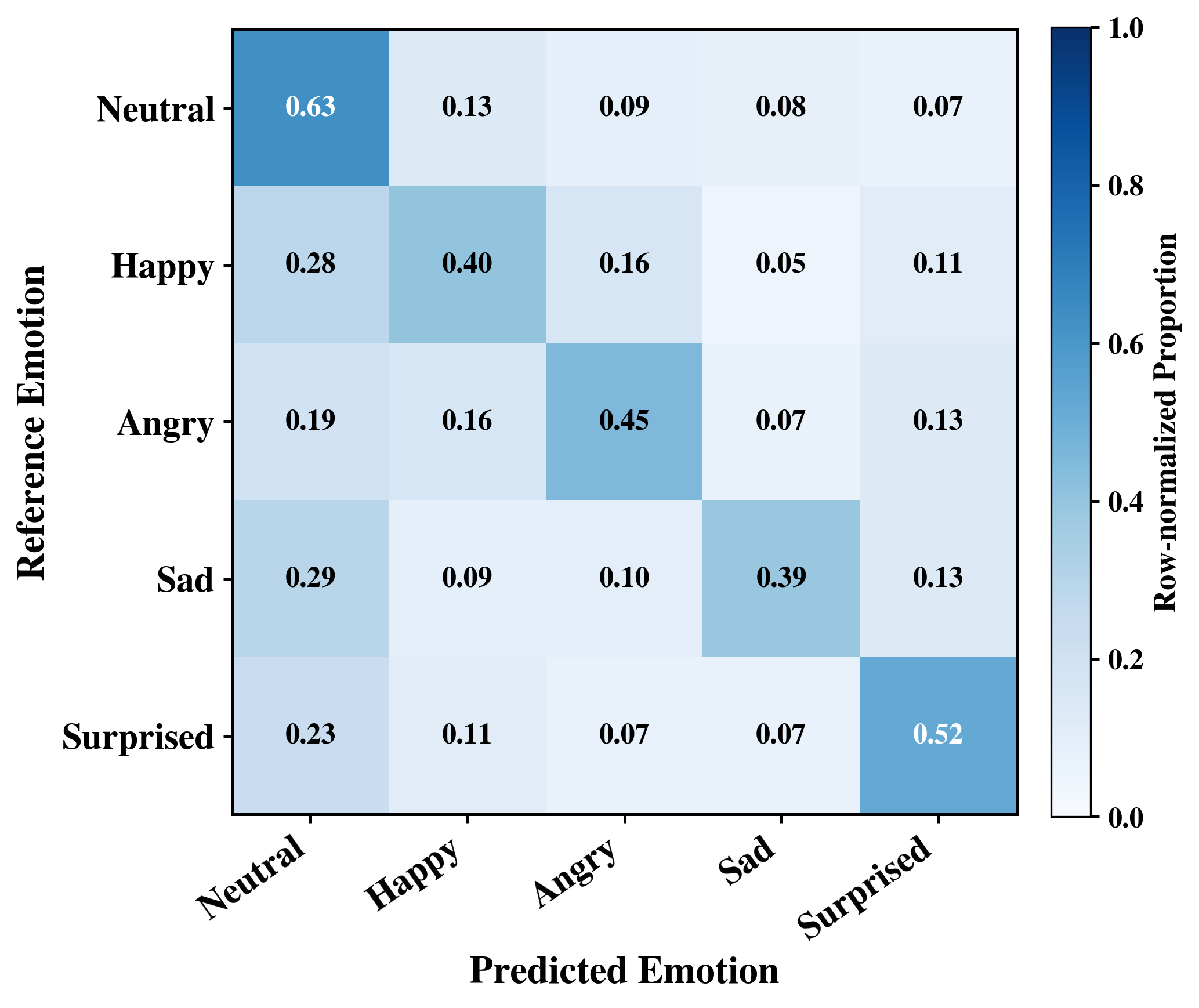}
    \caption{Chinese subset.}
    \end{subfigure}
}{%
    \begin{subfigure}[t]{0.47\textwidth}
    \centering
    \fbox{\rule{0pt}{1.8in}\rule{0.90\linewidth}{0pt}}
    \caption{English subset.}
    \end{subfigure}\hfill
    \begin{subfigure}[t]{0.47\textwidth}
    \centering
    \fbox{\rule{0pt}{1.8in}\rule{0.90\linewidth}{0pt}}
    \caption{Chinese subset.}
    \end{subfigure}
}
\caption{Row-normalized confusion matrices on the English and Chinese subsets of \mepa. Because each lexical group shares the same statement across five emotional realizations, off-diagonal mass directly reflects failures of acoustic grounding under controlled lexical content.}
\Description{A pair of row-normalized confusion matrices for the English and Chinese subsets of MEPA.}
\label{fig:mepa_confusion}
\end{figure*}

\subsection{Main Results}
\label{subsec:main_results}

\paragraph{SER Comparison.}

We first ask whether explicit para-semantic factorization improves standard SER not only on familiar corpora but also under cross-corpus and cross-lingual shift. Table~\ref{tab:main_results} shows that \model attains the highest average accuracy across the full nine-dataset suite, reaching 72.10\%. The improvement is not concentrated on a single benchmark. Instead, \model ranks first on MELD, RAVDESS, SAVEE, MERBench Test1, MERBench Test2, and AESDD, indicating that the same representation strategy remains effective across conversational, acted, and multilingual zero-shot settings.

This comparison also clarifies the complementary strengths of the strongest baselines. EmotionThinker remains strongest on IEMOCAP, while BLSP-Emo performs best on RESD and CaFE. \model does not dominate every corpus individually, yet it exhibits the most favorable aggregate profile. This pattern matters because the central goal is not to overfit one dataset family, but to maintain emotion sensitivity without sacrificing the lexical stability required for speech-conditioned language modeling. The gains therefore support the claim that a dedicated paralinguistic pathway improves robustness when prosody, recording style, and language vary simultaneously.

\paragraph{Empathetic Response.}
\begin{table}[t]
\centering
\small
\caption{SER accuracy and empathetic response results on SpeechAlpaca.}
\label{tab:speechalpaca_results}
\setlength{\tabcolsep}{4pt}
\begin{tabular}{lccc}
\toprule
\textbf{Method} & \textbf{SER $\uparrow$} & \textbf{Quality $\uparrow$} & \textbf{Empathy $\uparrow$} \\
\midrule
BLSP \cite{wang2024blsp} & 36.8 & 5.41 & 4.72 \\
Qwen2-Audio \cite{chu2024qwen2audio} & 52.6 & 5.88 & 5.03 \\
MiniCPM-O \cite{openbmb2025minicpmo} & 46.2 & 6.02 & 5.16 \\
OSUM-EChat \cite{geng2025osumechat} & 56.7 & 6.67 & 5.74 \\
BLSP-Emo \cite{wang2024blspemo} & \underline{83.8} & \underline{6.90} & \underline{5.91} \\

\textbf{EMODE} & \textbf{88.75} & \textbf{7.12} & \textbf{6.14} \\
\bottomrule
\end{tabular}
\end{table}

We next evaluate whether improved affective representations translate into better response behavior. Table~\ref{tab:speechalpaca_results} reports performance on SpeechAlpaca, where the main objective is to produce a helpful and emotionally appropriate reply from speech alone. Under this protocol, \emph{quality} and \emph{empathy} are the primary metrics, whereas the auxiliary SER score serves only as a supporting reference. This distinction is important: a model may recover the correct emotion label yet still generate a reply that is generic, unhelpful, or poorly aligned with the speaker's affective state.

This contrast is visible in the comparison with BLSP-Emo. Although BLSP-Emo remains competitive on the auxiliary SER measure, \model achieves better results on both judge-based empathetic metrics. The result is consistent with the intended use of para-semantic factorization: emotion understanding is most valuable when it improves downstream response generation rather than remaining an isolated recognition capability.

\paragraph{Controlled Lexical Grounding.}
\begin{table}[t]
\centering
\small
\caption{MEPA benchmark results under controlled lexical content. GLCA denotes Group-Level Complete Accuracy, which requires all five emotional realizations of the same lexical statement to be predicted correctly.}
\label{tab:mepa_results}
\setlength{\tabcolsep}{2.1pt}
\begin{tabular}{lcccccc}
\toprule
\textbf{Model} & \multicolumn{3}{c}{\textbf{English}} & \multicolumn{3}{c}{\textbf{Chinese}} \\
\cmidrule(lr){2-4}\cmidrule(lr){5-7}
& \textbf{Acc.} & \textbf{Macro-F1} & \textbf{GLCA} & \textbf{Acc.} & \textbf{Macro-F1} & \textbf{GLCA} \\
\midrule
Qwen2-Audio \cite{chu2024qwen2audio} & 44.6 & 42.1 & 2.7 & 35.9 & 33.4 & 1.0 \\
BLSP-Emo \cite{wang2024blspemo} & 50.4 & 48.3 & 6.8 & 40.7 & 38.5 & 3.1 \\
EmotionThinker \cite{wang2026emotionthinker} & \underline{52.1} & \underline{50.2} & \underline{8.1} & \underline{43.0} & \underline{40.7} & \underline{4.9} \\

\textbf{EMODE} & \textbf{56.4} & \textbf{54.7} & \textbf{12.1} & \textbf{47.8} & \textbf{45.3} & \textbf{8.6} \\
\bottomrule
\end{tabular}
\end{table}

We then isolate a different question from the main SER comparison: does the model still rely on the correct acoustic evidence once lexical shortcuts are removed? MEPA is designed specifically for this purpose. Within each evaluation group, the lexical statement is fixed and only the speaking style changes across the five emotions. Standard sample-level accuracy and macro-F1 therefore remain useful, but GLCA is especially informative because it counts a group as correct only when all five emotional realizations are predicted correctly. The sizable gap between sample-level metrics and GLCA is therefore expected rather than anomalous; it reflects the difficulty of sustaining consistent acoustic grounding across a complete lexically controlled set.

Table~\ref{tab:mepa_results} shows that \model ranks first on both English and Chinese across all three metrics. The advantage is most pronounced on GLCA, where \model exceeds EmotionThinker by 4.0 points on English and 3.7 points on Chinese. This pattern indicates that the gain is not limited to single-utterance recognition. Once lexical content is controlled, \model also yields more stable group-level discrimination across the full five-emotion inventory, consistent with the intended functional separation of lexical and paralinguistic evidence.

Figure~\ref{fig:mepa_confusion} makes the same pattern visible at the error-distribution level. The English subset retains a clearer diagonal structure, particularly for \emph{happy}, \emph{angry}, and \emph{surprised}. By contrast, the Chinese subset exhibits a stronger drift toward \emph{neutral} and a broader spread of off-diagonal confusion. This error profile is consistent with the lower Chinese accuracy, macro-F1, and GLCA values in Table~\ref{tab:mepa_results}, indicating that acoustic grounding under fixed lexical content remains less stable in Chinese than in English.

The confusion structure also clarifies the role of MEPA alongside conventional SER benchmarks. On standard corpora, models may exploit lexical regularities correlated with emotion labels. MEPA reduces this cue by construction, so its results more directly test sensitivity to acoustic-prosodic variation. The stronger neutral attraction and wider off-diagonal dispersion in Chinese suggest less stable affective grounding than in English, rather than failure on only one emotion class. Because MEPA is synthetic, these results should be interpreted as a controlled stress test rather than a substitute for spontaneous speech evaluation.

\subsection{Ablation and Mechanistic Analysis}
\label{subsec:ablation_study}

The following analyses address three questions: how much each training objective contributes to the final SER profile, whether GDR stabilizes routing and paralinguistic specialization, and which fusion rule best recombines the semantic and paralinguistic streams.

\begin{table}[htbp]
\centering
\small
\caption{Objective ablation on four representative SER benchmarks. Results are accuracy (\%).}
\label{tab:ablation_objectives}
\setlength{\tabcolsep}{3pt}
\begin{tabular}{lccccc}
\toprule
\textbf{Variant} & \textbf{IEMOCAP} & \textbf{MELD} & \textbf{RAVDESS} & \textbf{SAVEE} & \textbf{Avg} \\
\midrule
EMODE & \textbf{76.68} & \textbf{64.23} & \textbf{75.58} & \textbf{85.76} & \textbf{75.56} \\
w/o oeg-sem & 74.52 & 62.31 & 72.94 & 82.18 & 72.99 \\
w/o oeg-para & 71.84 & 59.37 & 68.12 & 79.24 & 69.64 \\
w/o SAA & 74.86 & 62.88 & 71.95 & 82.94 & 73.16 \\
w/o GDR & 74.23 & 62.57 & 70.64 & 81.86 & 72.33 \\
\bottomrule
\end{tabular}
\end{table}

Table~\ref{tab:ablation_objectives} shows that the full objective set yields the strongest average SER accuracy. The largest degradation occurs without \textbf{oeg-para}, indicating that explicit affective supervision makes the largest contribution among the tested objectives. Removing \textbf{oeg-sem} also weakens performance, especially on RAVDESS and SAVEE, suggesting that lexical anchoring remains important even when the final task is emotion recognition. Removing \textbf{SAA} or \textbf{GDR} produces smaller but consistent declines across the four datasets. The pattern is therefore not attributable to a single auxiliary term: OEG provides branch-specific supervision, while SAA and GDR contribute complementary cross-modal alignment and routing regularization to the final objective set.

\begin{table}[htbp]
\centering
\small
\caption{Fusion design comparison on four representative SER benchmarks. Results are accuracy (\%).}
\label{tab:ablation_fusion}
\setlength{\tabcolsep}{3pt}
\begin{tabular}{lccccc}
\toprule
\textbf{Fusion design} & \textbf{IEMOCAP} & \textbf{MELD} & \textbf{RAVDESS} & \textbf{SAVEE} & \textbf{Avg} \\
\midrule
Residual para & \textbf{76.68} & \textbf{64.23} & \underline{75.58} & \textbf{85.76} & \textbf{75.56} \\
Scaled residual & \underline{76.11} & \underline{63.92} & \textbf{75.71} & \underline{85.34} & \underline{75.27} \\
Weighted sum & 75.47 & 62.94 & 74.62 & 84.48 & 74.38 \\
Concat MLP & 75.92 & 63.41 & 75.05 & 84.91 & 74.82 \\
\bottomrule
\end{tabular}
\end{table}

Table~\ref{tab:ablation_fusion} compares several fusion rules under the same training recipe. Residual para-semantic fusion performs best on average, although its advantage is modest: scaled residual fusion is close and slightly better on RAVDESS. The two residual variants preserve the semantic stream while adding a routed paralinguistic correction, matching the intended asymmetry between lexical anchoring and affective modulation. Weighted summation and concatenation remain competitive but are weaker on average. These results favor the residual design in this controlled comparison without showing that it must dominate on every corpus.

\subsection{Connector Comparison}
\label{subsec:connector_comparison}

\begin{table}[htbp]
\centering
\small
\caption{Connector comparison on four SER benchmarks. Results are accuracy (\%).}
\label{tab:adapter_comparison}
\setlength{\tabcolsep}{3pt}
\begin{tabular}{lccccc}
\toprule
\textbf{Connector} & \textbf{IEMOCAP} & \textbf{MELD} & \textbf{RAVDESS} & \textbf{SAVEE} & \textbf{Avg} \\
\midrule
Subsampler & 71.04 & 60.58 & 69.17 & 78.96 & 69.94 \\
CFormer & 74.32 & 62.45 & 72.88 & 82.61 & 73.07 \\

DPSE & \textbf{76.68} & \textbf{64.23} & \textbf{75.58} & \textbf{85.76} & \textbf{75.56} \\
\bottomrule
\end{tabular}
\end{table}

Table~\ref{tab:adapter_comparison} compares \dpse with alternative speech-language connectors under the same training and evaluation protocol. The subsampler is weakest, CFormer narrows the gap, and \dpse remains best on all four benchmarks. The subsampler provides no explicit mechanism for separating lexical and affective evidence, whereas CFormer offers a stronger learned connector without branch-specific supervision. Within this controlled comparison, DPSE's consistent advantage supports structured para-semantic factorization rather than connector capacity alone.

\section{Conclusion}
\label{sec:conclusion}

In this paper, we introduced \textbf{EMODE}, an emotion-aware speech-language model designed to address the limitations of entangled acoustic representations. Its DPSE module organizes semantic and paralinguistic processing into dedicated pathways and recombines them through residual fusion. The coordinated three-stage curriculum uses Orthogonal Expert Guidance (OEG), Semantic-to-Acoustic Alignment (SAA), and Gating Diversity Regularization (GDR) to establish complementary supervision, align affective context, and discourage premature routing collapse.

EMODE obtains the highest average accuracy among the compared systems across a nine-dataset SER suite and the highest model-judged quality and empathy scores on SpeechAlpaca. Results on the controlled \mepa benchmark provide further evidence that EMODE uses acoustic-prosodic information when lexical content is held fixed. The objective and connector ablations further associate the gains with complementary expert supervision and structured fusion. Together, these results support explicit para-semantic factorization for affect-sensitive spoken dialogue systems. Future work will examine human evaluation, spontaneous affective speech, synthesis-robust evaluation, and broader multilingual interactions.

\clearpage
\begin{acks}
This work was supported by the Mobile Information Networks National Science and Technology Major Project (2026ZD1305800), Zhejiang Province's Leading Talent Project in Science and Technology Innovation (2023R5204), and Ningbo Key Research and Development Program (2025Z082).
\end{acks}

\bibliographystyle{ACM-Reference-Format}
\balance
\bibliography{sample-base}

\end{document}